\documentclass[12pt,oneside,runningheads]{taima}
\usepackage{graphicx, float, epsfig, url}
\usepackage{subcaption}
\usepackage{lineno}
\usepackage{color}
\usepackage{epstopdf}
\newsavebox{\measurebox}
\usepackage{adjustbox}
\usepackage{wrapfig,booktabs}
\usepackage{array}
\newcommand{\PreserveBackslash}[1]{\let\temp=\\#1\let\\=\temp}
\newcolumntype{C}[1]{>{\PreserveBackslash\centering}p{#1}}
\newcolumntype{R}[1]{>{\PreserveBackslash\raggedleft}p{#1}}
\newcolumntype{L}[1]{>{\PreserveBackslash\raggedright}p{#1}}

\begin{document}

\renewcommand\thelinenumber{\color[rgb]{0.5,0.5,0.5}\normalfont\sffamily\scriptsize\arabic{linenumber}\color[rgb]{0,0,0}}
\renewcommand\makeLineNumber {\hss\thelinenumber\ \hspace{6mm} \rlap{\hskip\textwidth\ \hspace{6.5mm}\thelinenumber}}

\pagestyle{headings}
\mainmatter

% Titre de la communication
\title{\Large{FaceOcc: A Diverse, High-quality Face Occlusion Dataset for Human Face Extraction}}
% En-têtes des pages impaires (sauf la 1ère)
\titlerunning{FaceOcc: A Diverse, High-quality Face Occlusion Dataset for Human Face Extraction}
\author{Xiangnan YIN and Liming CHEN}
% En-têtes des pages paires
\authorrunning{X. YIN and L. CHEN}
% Adresse des auteurs séparées par un "\and"
\institute{
    Département Mathématique-Informatique\\
    École Centrale de Lyon, \\
    Laboratoire LIRIS,\\
    36 Av. Guy de Collongue, 69134 Écully, France.\\
    Tél: int + 33 4 72 18 60 00, Fax: int + 33 4 78 43 39 62\\
    \email{\{yin.xiangnan, liming.chen\}@ec-lyon.fr}
    % \and
    % École Nat. Sup. des Télécommunications de Bretagne, \\
    % Département Image et Traitement de l'Information, \\
    % Technopôle Brest-Iroise, CS 83818, 29285 Brest Cedex, France.\\
    % Tél: int+ 33 2 29 00 10 61, Fax: int+ 33 2 29 00 10 98\\
    % \email{gwenael.brunet@enst-bretagne.fr\vspace{1cm}} \\
}

\maketitle %

% \begin{abstract}

% Ce document décrit un certain nombre de règles typographiques à respecter lors de la rédaction des
% communications pour les ateliers TAIMA. L'objectif visé par ces consignes est de garantir une unité
% de présentation des actes qui contribuera à la qualité générale des ateliers.

% \vspace{0.2cm}%
% \textbf{Mots clés} Taima, Instructions, Typographie, Modèles.
% \end{abstract}
% \selectlanguage{english}
\begin{abstract}
Occlusions often occur in face images in the wild, troubling face-related tasks such as landmark detection, 3D reconstruction, and face recognition. It is beneficial to extract face regions from unconstrained face images accurately. However, current face segmentation datasets suffer from small data volumes, few occlusion types, low resolution, and imprecise annotation, limiting the performance of data-driven-based algorithms. This paper proposes a novel face occlusion dataset with manually labeled face occlusions from the CelebA-HQ and the internet. The occlusion types cover sunglasses, spectacles, hands, masks, scarfs, microphones, etc. To the best of our knowledge, it is by far the largest and most comprehensive face occlusion dataset. Combining it with the attribute mask in CelebAMask-HQ, we trained a straightforward face segmentation model but obtained SOTA performance, convincingly demonstrating the effectiveness of the proposed dataset.

\vspace{0.2cm}%

\end{abstract}
\section{Introduction}
Human face extraction is a sub-domain of image segmentation, which aims to locate pure facial regions in face images, excluding backgrounds and occlusions (e.g., hair, hands, glasses, masks, and other facial accessories). It has a wide range of applications in face-related tasks such as face alignment~\cite{jia2014structured,zhao2019joint,yang2015robust}, face image de-occlusion~\cite{cai2020semi,lee2020byeglassesgan}, 3D face reconstruction~\cite{deng2019accurate,li2021fit,tran2018extreme}, and face recognition~\cite{zhi2019face,nirkin2018face,gao2019face}. With the widespread use of deep learning, significant progress has been made in image segmentation. Nevertheless, existing face extraction algorithms are still not sufficiently accurate, especially on face images in the wild, where occlusions of arbitrary shapes and textures can present.  We argue that the main reason for this problem is the lack of a high-quality occlusion-aware face segmentation dataset. The dataset should satisfy the following characteristics: 1) large volume of data, 2) various occlusion types, 3) accurate annotation, 4) high resolution. Unfortunately, no current dataset fulfills all four of these criteria. 

\begin{figure}
\centering
\sbox{\measurebox}{%
  \begin{minipage}[b]{.45\textwidth}
  \subfloat
    [CelebAMask-HQ]
    {\label{fig:1a}\includegraphics[width=\textwidth]{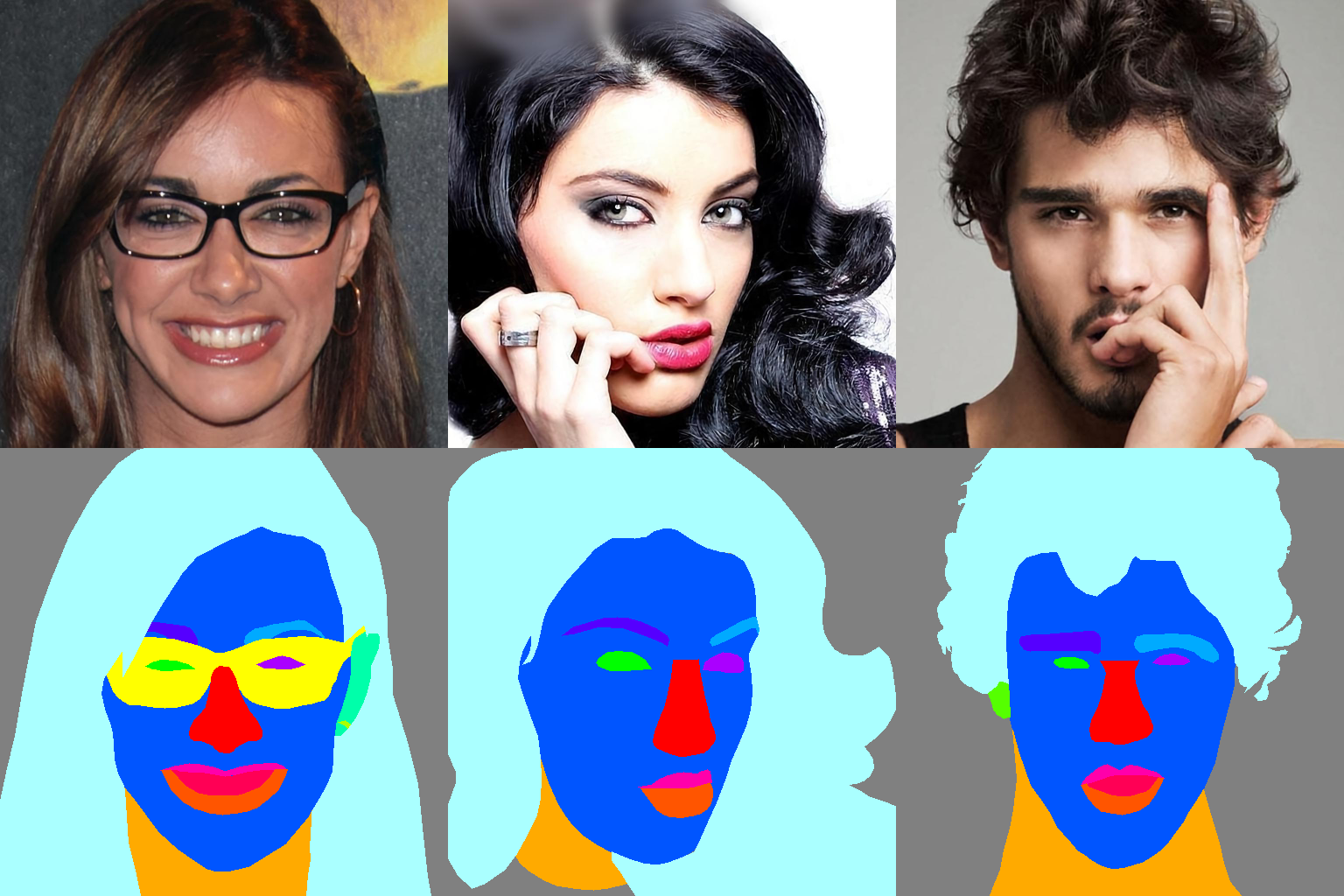}}
  \end{minipage}}
\usebox{\measurebox}\qquad
\begin{minipage}[b][\ht\measurebox][s]{.37\textwidth}
\centering
\subfloat
  [Part Label]
  {\label{fig:1b}\includegraphics[width=\textwidth]{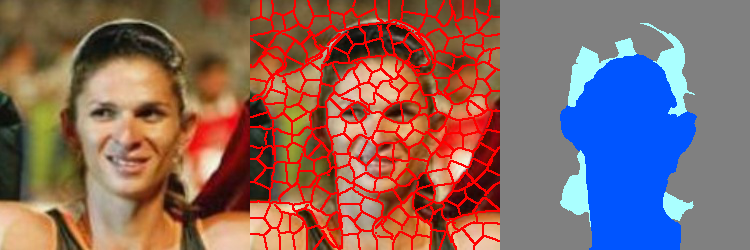}}
\vfill
\subfloat
  [HELEN]
  {\label{fig:1c}\includegraphics[width=\textwidth]{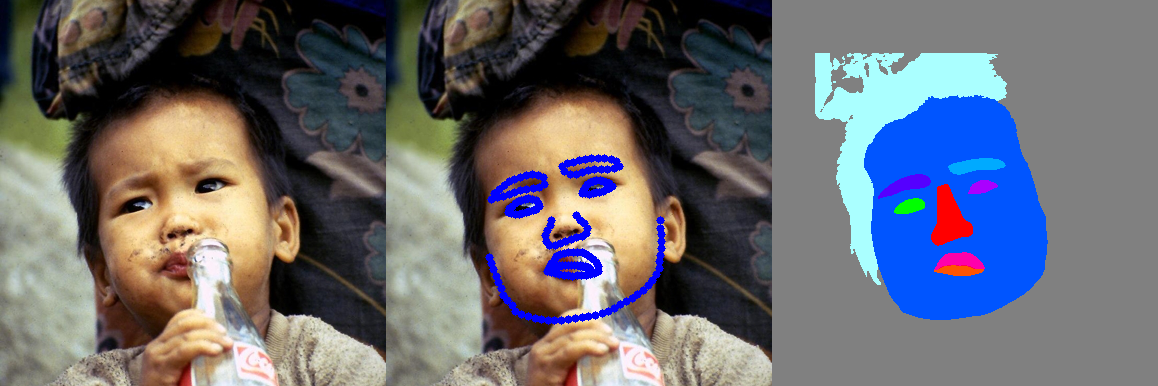}}
\end{minipage}
\caption{Problems of several existing datasets (cropped for better visualization).}
\label{fig:1}
\end{figure}

CelebAMask-HQ~\cite{CelebAMask-HQ} comes closest to meeting these requirements among the many face segmentation datasets. It provides manually annotated masks over 30K high-resolution face images in CelebA-HQ~\cite{karras2017progressive}, covering 18 facial attributes. Nevertheless, this dataset was designed for a GAN-based face image editing task, thus ignoring the various occlusions on the face image (except for hair and glasses), e.g., it wrongly annotates occlusions such as hands as attributes that they obscure. Furthermore, it classifies spectacles and sunglasses in the same category, which is unreasonable as the transparent lenses should not be considered an occlusion. Figure~\ref{fig:1a} illustrates the above problems. Despite the issues mentioned above, CelebAMask-HQ can be a good starting point for our work, significantly reducing the difficulty and effort of manual annotation. We examined 30K images in CelebA-HQ and annotated all occlusions ignored by CelebAMask-HQ, including hands, spectacles, microphones, scarfs, etc. As CelebA-HQ does not contain face images occluded by masks, we additionally collected such images using Google. Further, we also leverage numerous texture patches to replace the original textures of the occlusion, bringing a greater texture variety.

This paper describes how we construct the occlusion dataset and apply it to the face extraction task. And, we experimentally demonstrate the validity of the proposed method. Our contributions are as follows: 
\begin{itemize}
\item We propose a novel face occlusion dataset entitled FaceOcc, which is the largest and most comprehensive in this domain to the best of our knowledge. 
\item The proposed dataset, combined with the annotation of CelebAMask-HQ, allows generating a diverse range of augmented face masking data. 
\item A straightforward, lightweight face extraction model was trained on the proposed database, achieving SOTA performance without fancy metrics or model structures.
\end{itemize}

\section{Related Work}

Caltech Occluded Faces in the wild (COFW)~\cite{burgos2013robust} consists of 1007 heavily occluded face images with facial landmarks.  Although the dataset is designed for facial landmark detection under extreme conditions, given the large number of complex occlusions it contains, it has also been used in much of the literature~\cite{jia2014structured,ghiasi2015using,masi2020towards,nirkin2018face} for training and testing face extraction tasks. The biggest problem with this dataset is that its data volume is too small, with a training set of only 500 images, far from enough to train a deep neural network. Therefore many methods rely on other training data as a supplement, the most commonly used are Part Labels~\cite{GLOC_CVPR13} and HELEN~\cite{le2012interactive,smith2013exemplar}. 

Part Labels extends a subset of the Labeled Faces in the Wild (LFW)~\cite{Huang2007a} dataset by manually labeling each superpixel of the face images as hair, face, or background. Its annotation quality is poor since the superpixels provided by LFW are too coarse to capture the fine structure of the face and hair edges, as evidenced by Figure~\ref{fig:1b}. Based on manually labeled facial landmarks and a hair matting algorithm~\cite{levin2008spectral}, ~\cite{smith2013exemplar} annotated segmentation masks for 2 330 face images in the HELEN dataset. However, as shown in Figure~\ref{fig:1c}, the negligence of face occlusions and the unreliable hair matting algorithm resulted in coarse annotations similar to the Part Label. Recently, an updated version of the Part Labels, called Extended Labeled face in the wild (ELFW)~\cite{redondo2020extended}, has been proposed. It added new data, refined the existing segmentation maps, and synthesized large amounts of occluded face images. Despite its efforts in data volume, annotation quality, and diversity of occlusions, the improvements in all aspects are marginal. All of the above datasets are only in the order of thousands, which is still insufficient for data-driven approaches, and in addition, they are not aware of certain occlusions such as glasses. Table~\ref{tab:tab1} compares different aspects of the publicly available datasets of interest, including data volume, annotation quality, occlusion awareness, occlusion diversity, and whether data augmentation is applied.  

\begin{table}[t]
    \centering
    %\small
    %\begin{adjustbox}{width=\textwidth}
    \begin{tabular}{C{2.5cm} C{2cm} C{2cm} C{2cm} C{2cm} C{2cm}}
    \hline
    Name & Volume & Quality  & Occ Aware & Occ Diversity & Data Aug\\
    \hline
    COFW & 1007  & -- & -- & --  & No\\
    %\hline
    Part Labels & 2927 & Low & Partially & Low & No\\
    %\hline
    HELEN & 2330 & Low & Partially & Low & No\\
    ELFW & 3754 & Low & Partially & Low & Yes \\
    CelebAMask-HQ& 30 000 & High & Partially & High & No \\
    \hline
    \textbf{Ours} & \textbf{30 000}+ & \textbf{High} & \textbf{Yes} & \textbf{High} & \textbf{Yes} \\
    \hline
    \end{tabular}
    %\end{adjustbox}
    \vspace{2mm}
    \caption{Comparison of publicly available relevant datasets.}
    \label{tab:tab1}
    %\vspace{-10pt}
\end{table}

Some methods also annotated their private face segmentation datasets (of unknown quality) for training. ~\cite{saito2016real} segmented 5094 images from the FaceWarehouse~\cite{cao2013facewarehouse} dataset using GrabCut~\cite{rother2004grabcut} and manual inspection/correction. ~\cite{nirkin2018face} proposed a video-based semi-supervised face segmentation tool and generated 9818 segmented faces from 1275 videos in the IARPA Janus CS2 dataset~\cite{klare2015pushing}. Although they collected more training data, the labeling relies on scarce face videos, resulting in their dataset covering only 309 identities. We speculate that the dataset is of low diversity due to the videos' limited subjects and monotonous background region. ~\cite{masi2020towards} constructed a large face mask dataset with 598 266 images from CASIA-WebFaces~\cite{yi2014learning}, MS-Celeb-1M~\cite{guo2016ms}, and VGG Faces~\cite{parkhi2015deep}. Despite its large volume, the images are of low resolution, and the face masks derived from the pre-trained model of ~\cite{nirkin2018face} and the fitted 3D face silhouette are problematic. 

%As for data augmentation, besides the commonly used spatial and color transformations, the most representative technique is synthesizing occluded face images. ~\cite{nirkin2018face} generated occluded images with hand images and 3D sunglasses models. ELFW synthesized 2003 occluded face images with only 12 masks and 40 sunglasses. ~\cite{saito2016real} performed data augmentation with 1092 hand images and uniformly colored random-sized rectangles  (which is not reasonable), the former yielding 41 380 images and the latter 82 770 images. Our strengths over the above methods are that 1) we collected more face occlusions(3000+), 2) we built a texture dataset (800+) to make the synthesized occlusions rich in texture, 3) we perform data augmentation on the fly as training goes, instead of synthesizing fixed face images. 

\section{Proposed Method}
As mentioned above, CelebAMask-HQ is a large, high-quality face segmentation dataset with manually labeled facial attributes. Ideally, by integrating the attribute masks corresponding to the skin, eyes, nose, lips, and mouth, we obtain the mask of the whole face. However, the dataset ignores various face occlusions (except for hair and glasses) and the difference between spectacles and sunglasses, making this straightforward approach inaccurate. Consequently, we must label all of the occlusions in the dataset by hand to get an accurate face mask. This section describes how we construct the occlusion dataset and apply it to train our baseline model. 

%and perform a truncated subtraction between the above-integrated face mask and the occlusion mask

\subsection{Dataset Construction}
First, we aligned all face images in CelebA-HQ and resized them to a resolution of $256\times 256$ via facial landmarks detected by~\cite{guo2020towards}. This process ensures that the labeled face occlusions are of appropriate size and position for subsequent data augmentation. We then inspected the whole dataset and selected images with unlabeled or mislabeled occlusions. Next, we relabel those occlusions.

In many cases, the occlusion is not as evident and definitive as the hand or sunglasses, and different annotators may have different standards of judgment. Here list our annotation rules regarding several special occlusions: 
\begin{itemize}
    \item For spectacles with colorless lenses, only the reflections on the lenses (if present) and the frames are considered occlusions. 
    \item For tinted but transparent lenses, we determine them as skin if their color does not differ much from the skin, and vice versa for occlusion, but the labeling will bypass the eyes. 
    \item The tiny, heavy shadows cast by the eyeglass frames on the face are also labeled. 
    \item Makeups with a clear border with the skin are another source of occlusion.
    \item The tongue out of the mouth should be treated as occlusion. 
    \item All occlusions should be marked as a whole, not just the area above the face.
\end{itemize}

With the occlusion mask and the originally annotated attribute masks, we get the face mask by simple truncated subtraction:  
\begin{equation}
    M_{face} = \mathrm{max}(\hat{M}_{face}-M_{occ}, 0),
\end{equation}
where $\hat{M}_{face}$ denotes the integration of initially labeled attribute masks (including the face skin, eyes, nose, lips, and mouth), and $M_{occ}$ denotes the occlusion mask. 

Then we trained our first face extraction model on using $M_{face}$ and the images of CelebA-HQ (data augmentation method and model structure in the later sections). We selected hard samples from the FFHQ~\cite{karras2019style} with this model and added the labeled occlusions to our dataset. As neither CelebA-HQ nor FFHQ contains scarf and mask type occlusions, we collected those occlusions by Google to supplement our dataset.

All annotations were done with an Apple Pencil in the Magic Eraser~\footnote{https://apps.apple.com/us/app/magic-eraser-background-editor/id989920057}, and we ended up with a collection of over 3000 occlusions.

\subsection{Data Augmentation} %
\begin{figure}[t]
    \centering
    \includegraphics[width=0.9\linewidth]{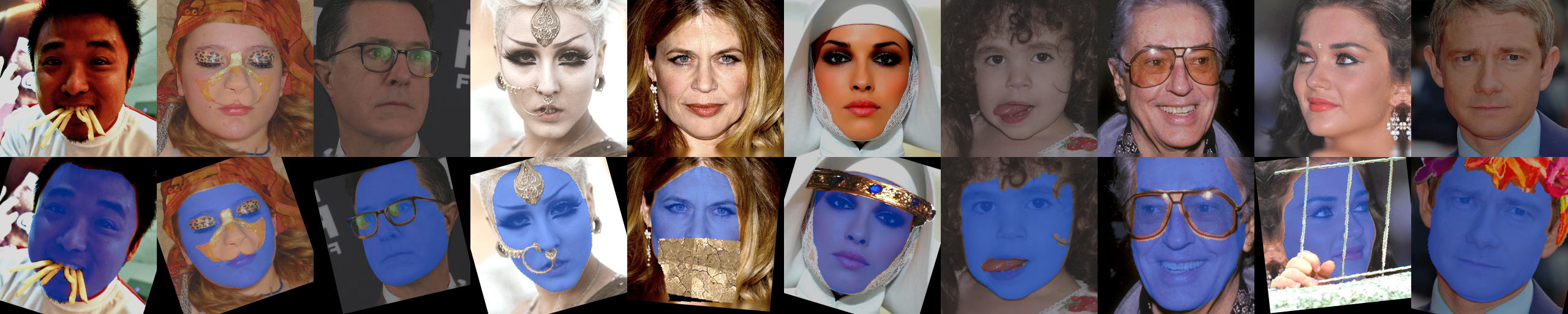}
    \caption{Results of annotation and data enhancement. The 1st row is original images; The 2nd row is augmented images with faces masked in blue.}
    \label{fig:my_label}
\end{figure}
We apply three data augmentation techniques: 1) spatial and color transformation of the face images, 2) randomly superimposing occlusions onto the face images, 3) randomly changing the texture of the occlusions (for masks and scarfs). Although some of the synthesized images are not realistic enough, experiments demonstrate that the improvement brought by our method is significant. Moreover, instead of synthesizing fixed face images, as many existing methods do~\cite{saito2016real,nirkin2018face,redondo2020extended}, we perform data augmentation on the fly as training goes, bringing more data diversity.

\subsection{Baseline Model}
Our baseline model uses ResNet-18~\cite{he2016deep} as the backbone and follows the U-Net~\cite{ronneberger2015u} structure. With the augmented data as input, the model predicts a single channel probability map representing the face mask. We employ the binary cross-entropy loss combined with ``online hard example mining" (OHEM)~\cite{shrivastava2016training} to guide the training. All the above settings are the most commonly used, without bells and whistles.   

\section{Experiments}
%https://tex.stackexchange.com/questions/258956/latex-wraptable-too-much-leading-space
\newlength{\oldintextsep}
\setlength{\oldintextsep}{\intextsep}
\setlength\intextsep{2pt}
\begin{wraptable}{l}{0pt}
\footnotesize
\begin{tabular}{L{2.6cm} C{.2cm} C{1cm} C{1cm} C{1cm} C{1cm}}
\toprule
Method & & IOU & Global & recall & FPS \\
\cline{1-1}
\cline{3-6}
Struct. Forest~\cite{jia2014structured} & & -- & 83.9 & -- & 88.6 \\
RPP~\cite{yang2015robust} & & 72.4 & -- & -- & 0.03 \\
SAPM~\cite{gao2019face} & & 83.5 & 88.6 & 87.1 & -- \\
Liu \textit{et al.}~\cite{liu2015multi} & & 72.9 & 79.8 & 89.9 & 0.29 \\
Saito \textit{et al.}~\cite{saito2016real} & & 83.9 & 88.7 & 92.7 & 43.2 \\
Nirkin \textit{et al.}~\cite{nirkin2018face} & & 83.7 & 88.8 & 94.1 & 48.6 \\
Masi \textit{et al.}~\cite{masi2020towards} & & 87.0 & 91.3 & 92.4 & \textbf{300} \\
\cline{1-1}
\textbf{Ours} & & \textbf{93.7} & \textbf{98.0} & \textbf{98.3} & 257\\
\bottomrule
\end{tabular}
\caption{COFW segmentation results. }\label{tab:result}
\end{wraptable}

We train our model for 30 epochs with a batch size of 16, which takes about 2 hours on two Nvidia GTX-1080 GPUs. The learning rate is initialized to $1e^{-4}$ and dropped to $1e^{-5}$ after the 20th epoch. The model is optimized using Adam optimizer with a weight decay of 0 and betas of (0.9, 0.999). We evaluate the model on 507 images of the COFW test set following the practice of previous works. Since none of the previous works published their annotations for this dataset, we manually re-labeled its test set. 

Table~\ref{tab:result} reports our quantitative results compared to others (data taken from the papers of Nirkin \textit{et al.} and Masi \textit{et al.}), where IOU refers to the intersection over union of the predicted mask and the ground truth, Global for the prediction accuracy of all pixels, recall for the percentage of correctly predicted face pixels to ground truth face pixels, and FPS for the number of images processed per second. Our method remarkably outperforms the others in the first three metrics and is slightly slower than Masi \textit{et al.}'s method. However, the input resolution of Masi \textit{et al.} is $128\times 128$, while ours is $256\times 256$.     

Experiment results show that the main factors limiting the performance of current face extraction models are the quality and quantity of training data. While using the proposed face occlusion dataset for training, even the most straightforward model can significantly outperform the state-of-the-art, strongly demonstrating the superiority of the our dataset. 

%a single GTX-1080 with a batch size of 1.  

\section {Conclusion}
This paper proposes a novel diverse, high-quality face occlusion dataset entitled FaceOcc, which contains all mislabeled occlusions in CelebAMask-HQ and complements some occlusions and textures from the internet. Together with the facial attribute masks in CelebAMask-HQ, the proposed dataset yields face masks and augmented data for training face extraction models. We validate the superiority of the proposed dataset by training a straightforward face extraction model far exceeding the state-of-the-art. 

%%%%%%%%%%%%%%%%%%%%%%%%%%%%%%%%%%%%%%%%%%%%%%%%%%%%%%%%%%%%%%%%%%%%%%%
% La bibliographie
%\nocite{*}
\bibliographystyle{plain}
\bibliography{InstructionsAuteurs}
% \clearpage\mbox{}Page \thepage\
% \clearpage\mbox{}Page \thepage\
% \clearpage\mbox{}Page \thepage\
% \clearpage\mbox{}Page \thepage\
%C'est la dernière page de la soumission.
\par\vfill\par
%La taille maximale d'une soumission TAIMA 2020 est 10 pages.
\end{document}